\documentclass[conference]{IEEEtran}
\ifCLASSINFOpdf
\else
\fi

\usepackage{amsmath,graphicx}
\usepackage{hyperref}
\usepackage{subcaption}
\usepackage{booktabs}

\begin{document}
%
% paper title
% Titles are generally capitalized except for words such as a, an, and, as,
% at, but, by, for, in, nor, of, on, or, the, to and up, which are usually
% not capitalized unless they are the first or last word of the title.
% Linebreaks \\ can be used within to get better formatting as desired.
% Do not put math or special symbols in the title.
\title{A Novel Dataset for Evaluating and Alleviating Domain Shift for Human Detection in Agricultural Fields}

% author names and affiliations
% use a multiple column layout for up to three different
% affiliations
\author{\IEEEauthorblockN{Paraskevi Nousi$^1$, Emmanouil Mpampis$^1$, Nikolaos Passalis$^1$, Ole Green$^2$, Anastasios Tefas$^1$}
\IEEEauthorblockA{$^1$Computational Intelligence and Deep Learning Group, Department of Informatics,\\
Aristotle University of Thessaloniki, Thessaloniki, Greece\\
$^2$Agrointelli, Aarhus, Denmark \\
E-mails: \{paranous,empampis,passalis\}@csd.auth.gr, olg@agrointelli.com, tefas@csd.auth.gr
}}

% conference papers do not typically use \thanks and this command
% is locked out in conference mode. If really needed, such as for
% the acknowledgment of grants, issue a \IEEEoverridecommandlockouts
% after \documentclass

% for over three affiliations, or if they all won't fit within the width
% of the page, use this alternative format:
% 
%\author{\IEEEauthorblockN{Michael Shell\IEEEauthorrefmark{1},
%Homer Simpson\IEEEauthorrefmark{2},
%James Kirk\IEEEauthorrefmark{3}, 
%Montgomery Scott\IEEEauthorrefmark{3} and
%Eldon Tyrell\IEEEauthorrefmark{4}}
%\IEEEauthorblockA{\IEEEauthorrefmark{1}School of Electrical and Computer Engineering\\
%Georgia Institute of Technology,
%Atlanta, Georgia 30332--0250\\ Email: see http://www.michaelshell.org/contact.html}
%\IEEEauthorblockA{\IEEEauthorrefmark{2}Twentieth Century Fox, Springfield, USA\\
%Email: homer@thesimpsons.com}
%\IEEEauthorblockA{\IEEEauthorrefmark{3}Starfleet Academy, San Francisco, California 96678-2391\\
%Telephone: (800) 555--1212, Fax: (888) 555--1212}
%\IEEEauthorblockA{\IEEEauthorrefmark{4}Tyrell Inc., 123 Replicant Street, Los Angeles, California 90210--4321}}

% use for special paper notices
%\IEEEspecialpapernotice{(Invited Paper)}

% make the title area
\maketitle

% As a general rule, do not put math, special symbols or citations
% in the abstract
\begin{abstract}
In this paper we evaluate the impact of domain shift on human detection models trained on well known object detection datasets when deployed on data outside the distribution of the training set, as well as propose methods to alleviate such phenomena based on the available annotations from the target domain. Specifically, we introduce the OpenDR Humans in Field dataset, collected in the context of agricultural robotics applications, using the Robotti platform, allowing for quantitatively measuring the impact of domain shift in such applications. Furthermore, we examine the importance of manual annotation by evaluating three distinct scenarios concerning the training data: a) only negative samples, i.e., no depicted humans, b) only positive samples, i.e., only images which contain humans, and c) both negative and positive samples. Our results indicate that good performance can be achieved even when using only negative samples, if additional consideration is given to the training process. We also find that positive samples increase performance especially in terms of better localization. The dataset is publicly available for download at \url{https://github.com/opendr-eu/datasets}.
\end{abstract}

% no keywords

% For peer review papers, you can put extra information on the cover
% page as needed:
% \ifCLASSOPTIONpeerreview
% \begin{center} \bfseries EDICS Category: 3-BBND \end{center}
% \fi
%
% For peerreview papers, this IEEEtran command inserts a page break and
% creates the second title. It will be ignored for other modes.
\IEEEpeerreviewmaketitle

\section{Introduction}
\label{sec:intro}

Object detection combines the tasks of classification and localization, i.e., it refers to finding \emph{what} objects are pictured in an image, as well as \emph{where} in the image they are located. In the case of multiple object, multiple class object detection, a generic object detector should be able to detect an unknown number of objects belonging to a number of different classes. Depending on the training dataset, these classes can include people, animals, inanimate objects, etc. Such datasets include the widely popular PASCAL VOC \cite{everingham2009voc} and MS COCO \cite{lin2014coco} object detection benchmarks, containing objects from 20 and 80 classes respectively. Deep Learning brought significant improvements both in terms of effectiveness and efficiency and currently the top performing object detection methods on these challenging benchmarks are all based on Deep Convolutional Neural Networks (CNNs). 

\begin{figure}[t!]
     \centering
     \begin{subfigure}[b]{0.4\textwidth}
         \centering
         \includegraphics[width=\textwidth]{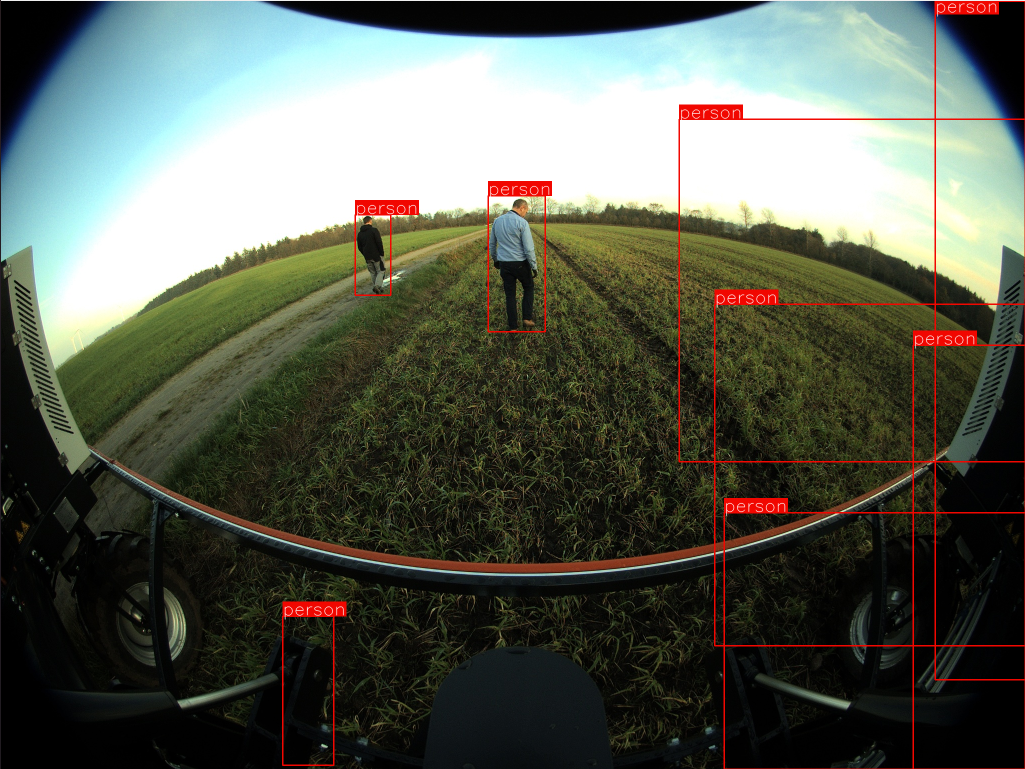}
         \caption{Image depicting two humans, captured by the front camera of the robot.}
         \label{fig:coco_two_humans}
     \end{subfigure}%
     \vspace{1em}\\
     \begin{subfigure}[b]{0.4\textwidth}
         \centering
         \includegraphics[width=\textwidth]{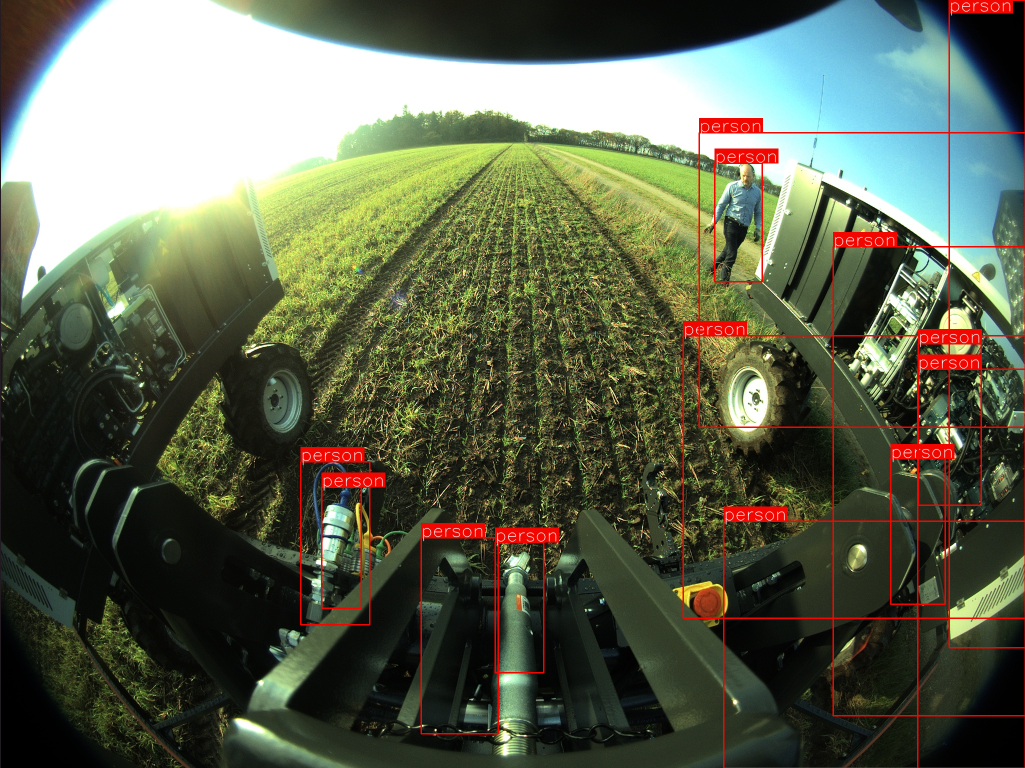}
         \caption{Image depicting one human, captured by the back camera of the robot.}
         \label{fig:coco_one_human}
     \end{subfigure}
     \caption{Examples of detections using a pretrained model. Existing models lead to false detections under distribution shifts.}
     \label{fig:baseline_detections}
\end{figure}

Despite improvements in these detectors on well known object detection benchmarks, deploying them on new applications highlights the domain adaptation problem. Domain adaptation refers to the process of alleviating the domain shift between a source and target domain, i.e., how to effectively deploy a detector trained on a source domain onto a target domain, which differs from the source in some way. Recent works in domain adaptation for object detection propose a progressive shift towards the target domain \cite{inoue2018cross,hsu2020progressive}. A somewhat more straightforward approach to domain adaptation is incremental learning \cite{wei2020incremental}. In any case, knowledge transfer is of significant importance when training a detector on a new dataset, in order to maximize its accuracy for the new domain. 

In this paper, we introduce a dataset, called OpenDR Humans in Field,  collected in the context of  agricultural use cases using the Robotti robotic platform, designed for detection of humans in fields. %Samples from this dataset are shown in Figure~\ref{fig:dataset_samples}.
Samples from this dataset are shown in Figure~\ref{fig:baseline_detections}, with detections made using an SSD detector \cite{liu2016ssd} pretrained on the COCO dataset \cite{lin2014coco}, where the domain shift problem is evident. We evaluate the accuracy of various detectors trained on existing datasets, identifying important limitations that these detectors face on scenarios like this. The results of this evaluation highlight the need to use domain adaptation and knowledge transfer approaches to increase the performance of detection on such applications. We examine various methods in depth and report results in terms of precision and recall for each. Our main findings are that using only negative samples can significantly drop the false positive rate, compared to a baseline pretrained model, and incorporating positive samples can further improve localization, leading to increased detection precision. 

\section{Related Work}
\label{sec:related}

Single-stage detectors have been shown to perform about as well as their two-stage, heavyweight counterparts, while running at much faster speeds. The seminal methods of YOLO \cite{redmon2018yolov3} and SSD \cite{liu2016ssd} inspired many recent works which utilize the anchor-based, single-stage architectures proposed by them. %Multiple variations of both methods have emerged, attempting to improve either the speed or accuracy, some of the most popular of which use different backbones to accommodate different tasks, like SSD Mobilenet \cite{huang2017speed}. 
Anchor-free object detectors aim to tackle issues arising from the use of predefined anchors, such as the need for thousands of such anchors in order to train dense object detectors, or the tedious hyperparameters they introduce, like the size, aspect ratio etc. CenterNet \cite{duan2019centernet} is one such anchor-free object detector, taking into consideration the center of objects as well as the corners, to detect each object as a triplet. 

In the context of agriculture, object detection methods can assist robots in their tasks in various ways \cite{vasconez2019human}. We are specifically interested in the human-centric scenario, where human labour is complemented with robots, focusing on human detection in fields, which is a critical safety aspect for human-robot interaction. The main contribution of this paper is the collection of a dataset that depicts humans in agricultural fields in various conditions. This is in contrast to the most commonly used person detection datasets, where humans are depicted in urban scenarios. Furthermore, the lenses attached to the robot are wide-angled, leading to bounding boxes of different proportions than those commonly seen in existing datasets. Therefore, the collected dataset allows for evaluating the impact of domain shift, as well as employing method for reducing its effect.

Indeed, this domain shift problem \cite{tommasi2016learning}, compared to existing datasets, is evident in Figure~\ref{fig:dataset_samples}, and is encountered in other computer vision tasks as well, such semantic segmentation \cite{sankaranarayanan2018learning} or concept detection \cite{luo2019taking}. In object detection, a two-level domain adaptation approach was introduced in \cite{chen2018domain} for Faster R-CNN, on an image-level as well as on an instance-level. In \cite{kim2019diversify}, a multidomain-invariant representation learning process was proposed, using adversarial learning. In this work, we tackle the domain shift problem in a data-driven manner, influenced also by the lack of a large collection of images, paving the way for developing methods that can work under the challenging settings that are often encountered in agricultural applications. %We focus on the training process of the detectors, to enhance detection performance on the target dataset, while maintaining the knowledge amassed by pretraining on the source dataset.

The rest of this paper is structured as follows. Section~\ref{sec:related} presents several works related to object detection and domain adaptation. The dataset collection process, as well as the  employed methods for alleviating domain shift are described in Section~\ref{sec:proposed}. The results of our experimental study are presented and analyzed in Section~\ref{sec:experiments}. Finally, Section~\ref{sec:conclusions} concludes our work and summarizes our findings.

\section{Proposed Method}
\label{sec:proposed}

\subsection{Dataset Collection}

\begin{figure}[t!]
     \centering
     \begin{subfigure}[b]{0.35\textwidth}
         \centering
         \includegraphics[width=\textwidth]{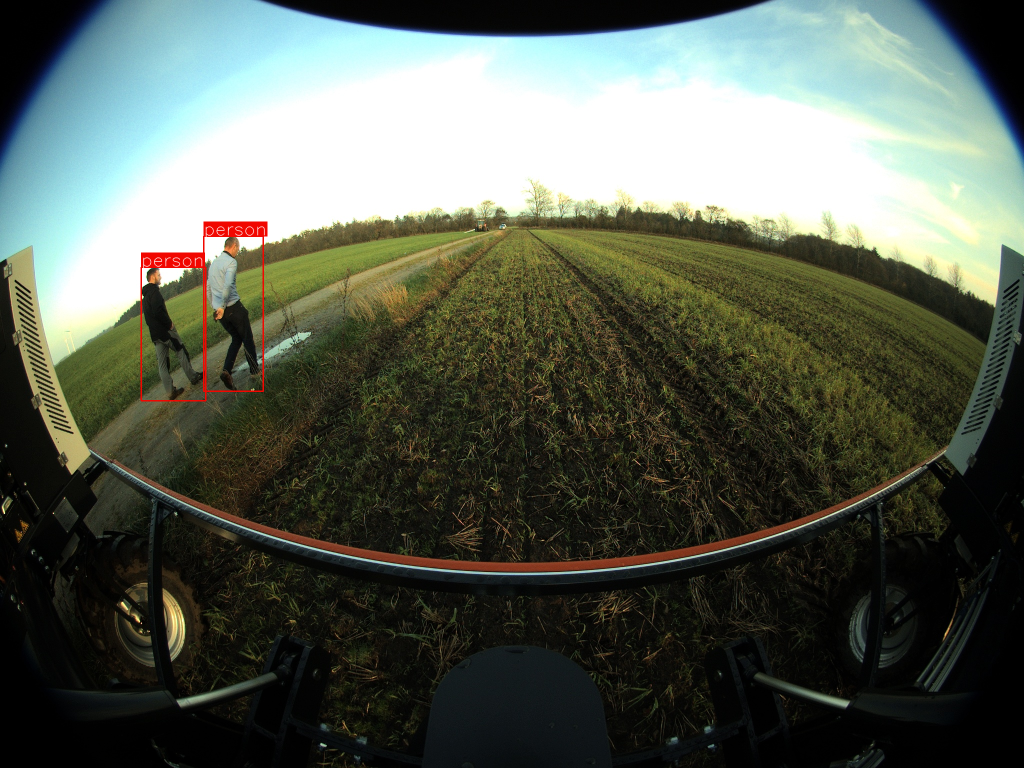}
         \caption{Positive sample depicting two humans.}
         \label{fig:human-dataset-annotated}
     \end{subfigure}%
     \vspace{1em}\\
     \begin{subfigure}[b]{0.35\textwidth}
         \centering
         \includegraphics[width=\textwidth]{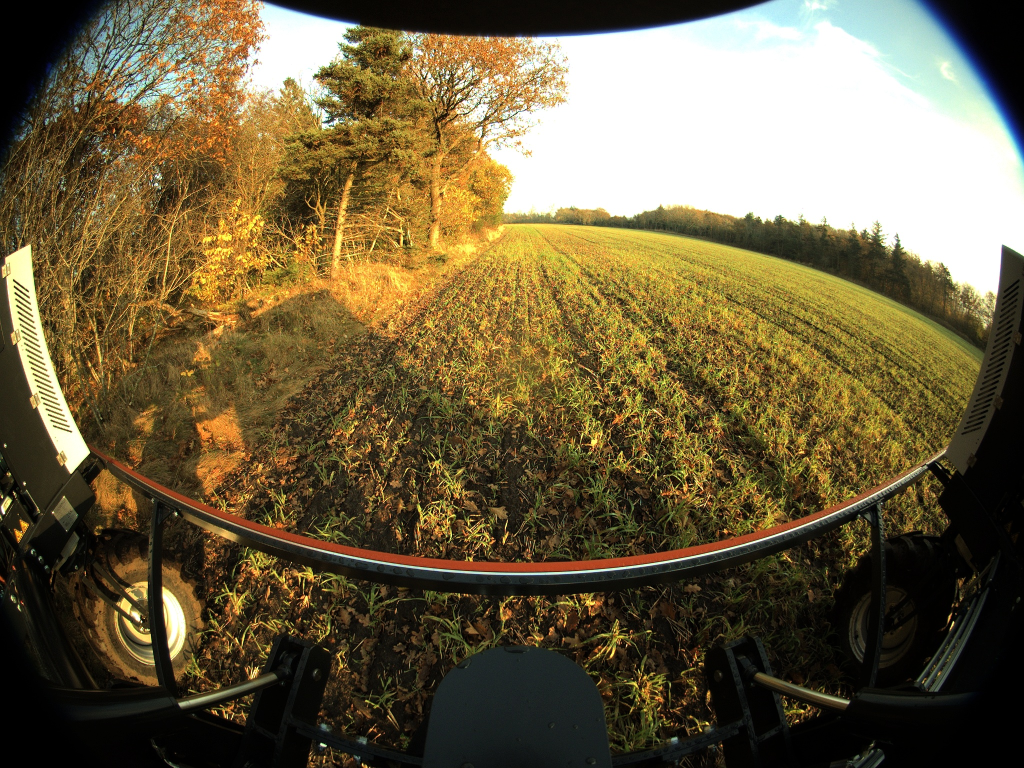}
         \caption{Negative sample with no humans in field of view.}
         \label{fig:human-dataset-no-human}
     \end{subfigure}
     \caption{Examples of collected images: (a) Image depicting two humans, annotated with bounding boxes, (b) Image depicting no humans.}
     \label{fig:dataset_samples}
\end{figure}

A Robotti was deployed by AGI to collect images with a front and back camera, in a realistic scenario to mimic the images that the robot might encounter in the agricultural use case. A total of 8038 images were collected on two separate occasions, 818 in the first batch and 7233 in the second. For the purposes of this work, the first batch was fully annotated, while the second one is provided to support unsupervised learning tasks.
Of the 818 collected images, 13 were discarded as they depicted unwilling participants to comply with GDPR. The remaining images were annotated with bounding boxes, where one bounding box corresponds to one depicted person. The LabelImg\footnote{\url{https://github.com/tzutalin/labelImg}} tool was used for the annotation, which outputs annotations in PASCAL VOC .xml format. Figure \ref{fig:human-dataset-annotated} is an image from this dataset annotated with two bounding boxes for the two depicted humans. In total, 158 images contained people, and 647 images did not. The latter were annotated with an empty bounding box list, to be used as negative samples in object detection algorithms. Figure \ref{fig:human-dataset-no-human} is an example of an image from this dataset containing no humans.

\subsection{Alleviating the Domain Shift}

A natural first step to alleviate the domain shift is to finetune a pretrained detector on background images from the new domain, i.e., images which do not depict any objects of interest. One benefit of this method is that no annotation is required, which can often be a tedious and time-consuming activity. However, training a detector solely on negative samples may quickly degrade the detector's performance on positive samples, i.e., images depicting objects of interest. A small learning rate and only a small number of training iterations can be used to lessen this undesirable side-effect. 

In the absence of a sufficient number of negative training examples, or in the case where performance is still sub-par, positive samples should be included in the training set. Although bounding box annotation is costly, it is the most reliable way to increase detection performance on a new dataset. Positive samples can be used as the only training set as they contain some of the background information as well. Furthermore, finetuning a pretrained detector with both positive and negative samples can lead to fewer false positive detections in comparison to training with only positive samples, as well as fewer missed detections in comparison to using only negative samples. 

However, finetuning using only the target dataset can deteriorate the detector's performance on the source dataset due to catastrophic forgetting phenomena~\cite{kemker2018measuring}. Even though the source dataset may no longer be relevant, this drop in performance can be reflected in the target dataset, in the form of overfitting. Thus, we propose that in addition to the aforementioned training sets, the source dataset is used in finetuning as well. Specifically for human detection, only the `person' class from the source dataset is extracted and appended to the training set. To enforce balance between the samples of the source and target datasets, the samples of the latter are repeated multiple times. Each sample undergoes transformations, according to the data augmentation protocol of the detector, such that even if the same image is used twice in a batch, the detector sees a slightly different version of it.

\section{Experimental Results}
\label{sec:experiments}

\subsection{Baseline models}

An extensive evaluation of pretrained detectors of the SSD \cite{liu2016ssd}, YOLOv3 \cite{redmon2018yolov3} and CenterNet \cite{duan2019centernet} families, was conducted for this dataset. The results are summarized in Table~\ref{tab:human-dataset-ap} in terms of precision at 0.5 IoU and FPS on Jetson AGX. The detectors are trained on either the PASCAL VOC \cite{everingham2009voc} and MS COCO \cite{lin2014coco} object detection benchmarks, containing objects of 20 and 80 classes respectively. Finally, the MobileNet version of SSD \cite{huang2017speed,nousi2018convolutional} is also evaluated. For the target dataset, we evaluate the methods on two subsets: a) on the positive samples only (`Pos. Only'), and b) on the entire test set (`All'), including both positive and negative samples. The reason behind this choice is to examine the effect of each training method on the false positive detections.

\begin{table}[]
    \caption{Evaluation in terms of precision at 0.5 IoU, of pretrained detectors on the collected dataset depicting humans in field.}
    \label{tab:human-dataset-ap}
    \centering
    \begin{tabular}{ccccc}
    \toprule
        \textbf{Method} & \textbf{Train Set} & \textbf{Pos. Only} & \textbf{All} & \textbf{FPS} \\ \midrule
        % SSD & WIDER & 43.7 & 26.2 & 23.6 \\
        SSD & VOC & 53.5 & 42.2 & 23.6 \\
        SSD & COCO & 80.3 & 70.1 & 23.6 \\ %\addlinespace
        SSD - MBNet & VOC & 40.8 & 18.8 & 35 \\
        SSD - MBNet & COCO & 60.7 & 42.3 & 35 \\ %\addlinespace
        CenterNet & VOC & 43.4 & 28.6 & 16.1 \\
        CenterNet & COCO & 63.1 & 54.8 & 16.1 \\ %\addlinespace
        YOLOv3 & VOC & 63.4 & 60.9 & 15.2 \\ 
        YOLOv3 & COCO & 78.9 & 74.7 & 15.2 \\
        \bottomrule
    \end{tabular}
\end{table}

As expected, the addition of images without people highlights the false positive accumulation, due to the unseen backgrounds present in the dataset. Detectors trained on COCO seem to perform significantly better than those trained on VOC, which can be attributed to the wider range of appearance in people in the larger COCO dataset. The object scale in COCO is also more varied, containing people as small as 10 pixels in height. The YOLOv3 detector in general performs the best, but is the slowest of the evaluated detectors on the Jetson AGX. The SSD MobileNet variant, especially when trained on COCO, seems to give off the best speed/accuracy trade-off. Even so, the drop in precision is significant when considering negative-only samples.

Based on this experimental study, we conclude that further training of the detectors is necessary to avoid false positive detections as well as to increase the true positive ratio. Knowledge transfer from the COCO dataset seems to be the most promising direction, as it leads to the best precision for all detectors. Furthermore, we choose the SSD algorithm as it is the fastest of the compared ones, and specifically the standard VGG16 version, as it still runs at about real-time on the AGX while achieving higher performance than its MobileNet counterpart.

\subsection{Domain Adaptation Experiments}

Two major sets of experiments are conducted.  In the first case, the detector is finetuned using only the target dataset, and specifically different splits of it.  In the second case, the detector is finetuned using the target dataset as well as the COCO `person' subset, i.e., any images from the COCO dataset which depict humans. 

\subsubsection{Finetuning on target domain only}
%\noindent \textbf{Finetuning on target domain only.} %We evaluate the proposed domain adaptation methods on the target dataset as well as the COCO `person' subset. 
For the following experiments, we also measure the performance in terms of Average Precision (AP), Precision at 0.5 IoU threshold, and Recall at 0.5 IoU. Thus we can draw conclusions regarding the false positive (FP), false negative (FN) and localization performance of each method. Table~\ref{tab:posonly_results} contains the results of our study on the positive subset of the target dataset. In comparison to the pretrained model on COCO, using only negative samples increases all metrics, although it has the least significant effect on recall. This can be attributed to a very small change of the FN detections, i.e., the detector is only slightly better at finding humans it didn't before finetuning. The most significant change is in terms of Precision@0.5, translating to a smaller FP rate, i.e., the detector has learned to not make false predictions, as expected. In terms of AP, the increase is not as large, indicating that although the overall FP rate has improved, localization issues ensue. This highlights the need to add positive samples to the training set. 

On the other hand, using only positive samples, significantly increases the detection performance in terms of both precision and recall. Using both positive and negative samples further increases the precision at 0.5 IoU, at the cost of slightly worsened localization at higher thresholds. Furthermore, the effect on FN is negligible. This indicates that using positive only samples provides the detector with enough background (i.e., negative) samples to reach this peak performance. 

\begin{table}[t!]
    \caption{Target domain finetuning - Evaluation using only the positive samples of the dataset}
    \label{tab:posonly_results}
    \centering
    \small
    \begin{tabular}{lccc}
    \toprule
        \textbf{Train Set} & \textbf{AP} & \textbf{Precicion}@0.5 & \textbf{Recall}@0.5 \\ \midrule
        Baseline (COCO) & 49.8 & 80.3 & 55.4 \\
        Finetune with negatives & 50.6 & 83.1 & 55.5 \\
        Finetune with positives & 57.0 & 90.1 & 63.2 \\
        Finetune with both & 56.6 & 91.1 & 63.3 \\
        \bottomrule
    \end{tabular}
\end{table}

% Table~\ref{tab:coco_person_results} provides a performance comparison of the proposed training methods in terms of their effect on the pretrained model, as measured on the `person' class of COCO before and after finetuning. Training with only negative samples leads to the smallest performance loss, specifically $1.3\%$. Positive only training drops performance by $2.7\%$, whereas adding both positive and negative drops the AP by $5\%$.

% \begin{table}[ht!]
%     \centering
%     \begin{tabular}{lc}
%     \toprule
%         \textbf{Train Set} & \textbf{AP} \\ \midrule
%         coco & 37.0 \\
%         negatives & 35.7 \\
%         positives & 34.3 \\
%         both & 32.0 \\
%         \bottomrule
%     \end{tabular}
%     \caption{Average precision, precision at 0.5 IoU and recall at 0.5 IoU for only the positive samples of the dataset.}
%     \label{tab:coco_person_results}
% \end{table}

The performance of the proposed methods on the entire test set (both positive and negative samples) is shown in Table~\ref{tab:both_results}. Note that splitting the test set like this only affects the precision scores, and not the recall. Thus, we focus on the precision scores, and specifically on the FP and localization performance.

\begin{table}[t!]
\caption{Target domain finetuning - Evaluation using the entire (positive and background) samples of the dataset}
\small
    \label{tab:both_results}
    \centering
    \begin{tabular}{lccc}
    \toprule
        \textbf{Train Set} & \textbf{AP} & \textbf{Precicion}@0.5 & \textbf{Recall}@0.5 \\ \midrule
        Baseline (COCO) & 44.8 & 70.1 & 55.4 \\
        Finetune with negatives & 48.6 & 78.9 & 55.5 \\
        Finetune with positives & 56.7 & 89.3 & 63.2 \\
        Finetune with both & 56.5 & 90.9 & 63.3 \\
        \bottomrule
    \end{tabular}
\end{table}

All of the evaluated methods lead to more FP detections, which is expected as these occur on the added negatives-only subset. Other than this drop, the results are similar to those regarding the positives-only subset. Specifically, using negative samples only improves the baseline performance and the effect is more prominent on this set (+$8.8\%$ precision at 0.5 IoU, in comparison to +$2.8\%$ in positives-only). 

Using only positive and using both positive and negative samples both significantly improve the performance over the baseline pretrained model, and actually more or less reach the same performance as when evaluating only on positive samples. This result is consistent with the fact that positive samples contain a superset of the information presented in negative samples that is relevant to the detection algorithm.

\subsubsection{Finetuning on target and related source domain class}

%\noindent \textbf{Finetuning on target and related source domain class.} 
Mixing the source and target domains in a balanced manner during training may intuitively increase the performance of the detector on the target dataset even more, while maintaining performance on the source domain. Table~\ref{tab:both_plus_coco} summarizes the results of this experiment on the entire target dataset, i.e., the results are comparable to those in Table~\ref{tab:both_results}. The `COCO AP' column shows the AP on the person subset of COCO, to highlight performance loss on the source domain.

\begin{table}[t!]
    \caption{Finetuning using both the target and source domain - Evaluation using the entire (positive and background) samples of the dataset}
    \label{tab:both_plus_coco}
    \centering
    \small
    \begin{tabular}{lccc|c}
    \toprule
        \textbf{Train Set} & \textbf{AP} & \textbf{Prec}@0.5 & \textbf{Rec}@0.5 & \textbf{COCO AP} \\ \midrule
        Baseline (COCO) & 44.8 & 70.1 & 55.4 & 37.0 \\ 
        % neg & 48.6 & 78.9 & 55.5 & 35.7 \\
        COCO+Neg. & 49.1 & 79.4 & 53.5 & 36.4 \\ 
        % pos & 56.7 & 89.3 & 63.2 & 34.3\\
        COCO+Pos. & 61.9 & 94.3 & 67.0 & 34.9 \\ 
        % both & 56.5 & 90.9 & 63.3 & 32.0 \\
        COCO+Both & 60.7 & 93.9 & 70.1 & 36.8 \\
        \bottomrule
    \end{tabular}

\end{table}

The `COC+Neg./Pos./Both` entry indicates that the detector has been finetuned using the COCO `person' subset and the negative/positive/all samples of the target dataset. First, training with negative only samples, increases the precision in comparison to both the pretrained model on COCO, as well as the finetuned models reported in Table~\ref{tab:both_results}. Furthermore, training with positive samples significantly improves both the precision and recall scores, at the cost of $2.1\%$ AP in the person class of COCO. Adding both positive and negative samples preserves the most pre-existing knowledge, as indicated by the small loss in person AP, as well as the $3.1\%$ improvement in recall in the AGI dataset, in comparison to using only positive samples. These results indicate that combining source and target domain can always increase the precision compared to using data only from the target domain, as well as minimize the impact of catastrophic forgetting phenomena.

\subsection{Qualitative Results}

\begin{figure}[b!]
     \centering
     \begin{subfigure}[b]{0.37\textwidth}
         \centering
         \includegraphics[width=\textwidth]{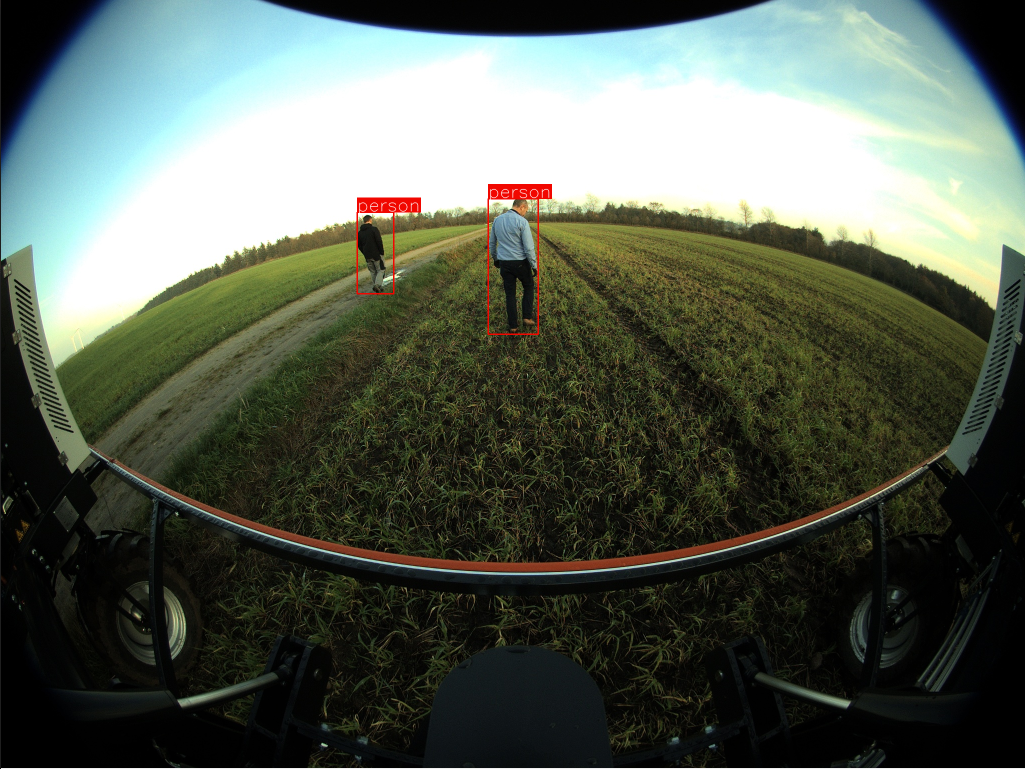}
         \caption{Positive sample depicting two humans (front camera).}
         \label{fig:our_two_humans}
     \end{subfigure}%
     \vspace{1em}\\
     \begin{subfigure}[b]{0.37\textwidth}
         \centering
         \includegraphics[width=\textwidth]{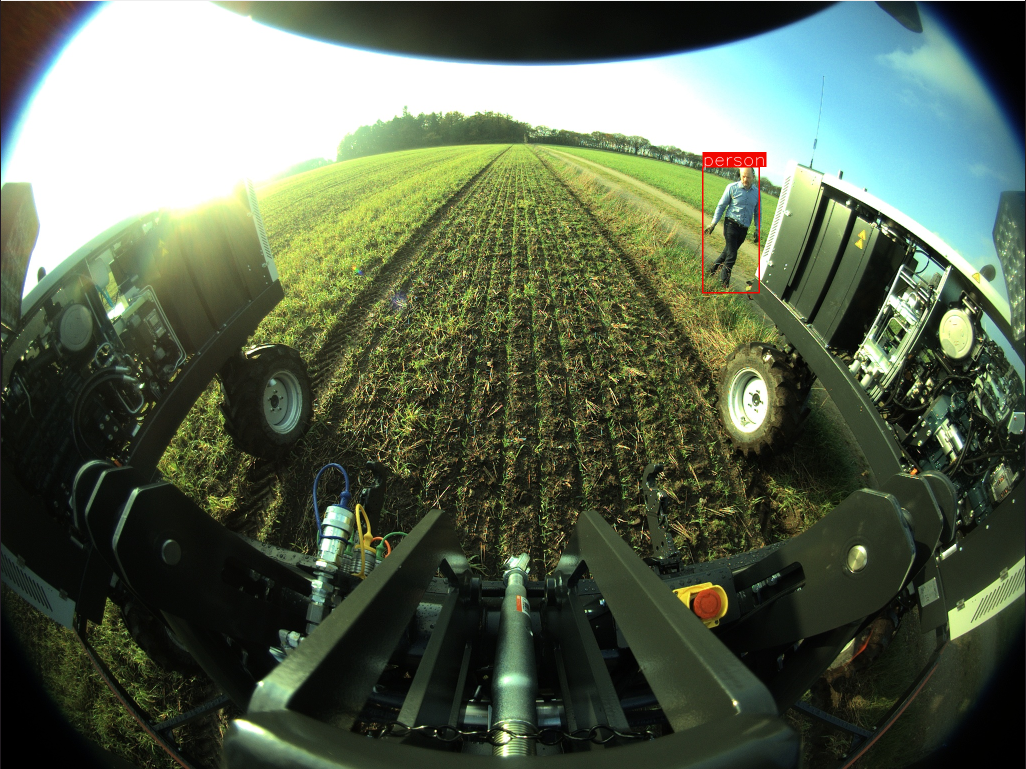}
         \caption{Positive sample depicting one human (back camera).}
         \label{fig:our_one_human}
     \end{subfigure}
     \caption{Examples of detections using our COCO+Both model: (a) Image depicting two humans, captured by the front camera of the robot, (b) Image depicting one human, captured by the back camera of the robot.}
     \label{fig:our_detections}
\end{figure}

Figure \ref{fig:our_detections} shows examples of detections made using our COCO+Both detector, on the same images as shown in Figure~\ref{fig:baseline_detections} for the pretrained model. Note that there are a lot of false positive detections using the pretrained model, which are corrected when training with the source (COCO) dataset plus the full annotated target domain dataset. 

\section{Conclusions}
\label{sec:conclusions}

A dataset for human detection in fields was introduced, for the purposes of agricultural robotics applications. Various detection models pretrained on the VOC and COCO datasets were evaluated on this dataset, and the results indicated a severe impact of the domain shift problem. Thus, the importance of annotation of the collected images was examined, by evaluating three distinct sets of training data: a) only negative samples, i.e., no depicted humans, b) only positive samples, i.e., only images which depict humans, and c) both negative and positive. The results indicated that good performance can be achieved even when using only negative samples. However, to achieve better localization, using positive samples only is the better option. %Finally, if the constraint of precise localization can be relaxed, using both negative and positive samples yields the best results.
The findings of this work, along with the openly available annotated dataset,  pave the way for developing methods that can work under the challenging settings that are often encountered in agricultural applications.

% conference papers do not normally have an appendix

% use section* for acknowledgment
\section*{Acknowledgment}
This project has received funding from the European Union’s Horizon 2020 research and innovation programme under grant agreement No 871449 (OpenDR). This publication reflects the authors views only. The European Commission
is not responsible for any use that may be made of the information it contains.

% trigger a \newpage just before the given reference
% number - used to balance the columns on the last page
% adjust value as needed - may need to be readjusted if
% the document is modified later
%\IEEEtriggeratref{8}
% The "triggered" command can be changed if desired:
%\IEEEtriggercmd{\enlargethispage{-5in}}

% references section

% can use a bibliography generated by BibTeX as a .bbl file
% BibTeX documentation can be easily obtained at:
% http://mirror.ctan.org/biblio/bibtex/contrib/doc/
% The IEEEtran BibTeX style support page is at:
% http://www.michaelshell.org/tex/ieeetran/bibtex/
\bibliographystyle{IEEEtran}
% argument is your BibTeX string definitions and bibliography database(s)
\bibliography{IEEEabrv,refs}
%
% <OR> manually copy in the resultant .bbl file
% set second argument of \begin to the number of references
% (used to reserve space for the reference number labels box)
%\begin{thebibliography}{1}

%\bibitem{IEEEhowto:kopka}
%H.~Kopka and P.~W. Daly, \emph{A Guide to \LaTeX}, 3rd~ed.\hskip 1em plus
%  0.5em minus 0.4em\relax Harlow, England: Addison-Wesley, 1999.
%
%\end{thebibliography}

% that's all folks
\end{document}